\documentclass{article}

\usepackage[preprint,nonatbib]{neurips_2018}

\usepackage{graphicx}
\usepackage{enumitem}   

\usepackage{caption}
\usepackage{subcaption}

\usepackage[utf8]{inputenc} 
\usepackage[T1]{fontenc}    
\usepackage{hyperref}       
\usepackage{url}            
\usepackage{booktabs}       
\usepackage{amsfonts}       
\usepackage{amsmath}
\usepackage{amssymb}
\usepackage{nicefrac}       
\usepackage{microtype}      
\usepackage{lipsum}
\usepackage{multicol}

\newcommand{\mb}{\mathbf}
\newcommand{\tb}{\textbf}

\title{Feature Selection Tutorial \\ with Python Examples
}

\author{
  P\'{a}draig Cunningham \\
  School of Computer Science\\
  University College Dublin\\
  \texttt{padraig.cunningham@ucd.ie} \\
   \And
   Bahavathy Kathirgamanathan \\
    School of Computer Science\\
   University College Dublin\\
   \texttt{bahavathy.kathirgamanathan@ucdconnect.ie}\\
   \And
 Sarah Jane Delany \\
  School of Computer Science\\
  Technological University Dublin\\
  \texttt{sarahjane.delany@tudublin.ie} \\
 }
 
\begin{document}
\maketitle

\begin{abstract}
In Machine Learning, feature selection entails selecting a subset of the available features in a dataset to use for model development. There are many motivations for feature selection, it may result in better models, it may provide insight into the data and it may deliver economies in data gathering or data processing. For these reasons feature selection has received a lot of attention in data analytics research. In this paper we provide an overview of the main methods and present practical examples with Python implementations. While the main focus is on supervised feature selection techniques, we also cover some feature transformation methods. 
\end{abstract}


\section{Introduction}
In data analysis, objects described using multiple features may sometimes be described using a subset of these features without loss of information. Identifying these feature subsets is termed feature selection, variable selection or feature subset selection and is a key process in data analysis. This paper provides a tutorial introduction to the main feature selection methods used in Machine Learning (ML). While excellent reviews and evaluations of feature selection methods already exist \cite{guyon2003introduction,molina2002feature} our main contribution is to provide examples of these methods in operation along with links to Python notebooks that implement these methods. 

Feature selection is important because it can deliver a number of benefits:

\begin{itemize}
    \item  \textbf{Better classifiers:} The obvious benefit of feature selection is that it will improve accuracy because redundant or noisy features can damage accuracy. Perhaps surprisingly, improvements in accuracy can be quite limited because powerful ML techniques are designed to be robust against noise. 
    \item \textbf{Knowledge discovery:}  Perhaps the most enduring benefit of feature selection is the insight it provides. Identifying influential features and features that are not useful teaches us a lot about the data. 
    \item \textbf{Data Gathering:}
    In domains where data comes at a cost (e.g. Medical Diagnosis, Manufacturing), identifying a minimal set of features for a classification task can save money. 
    \item \textbf{Computational Cost:} Identifying minimal feature subsets will allow for simpler models that will cost less to set up and run. 
    \item \textbf{The Curse of Dimensionality:} In theory, according to the Curse of Dimensionality, the amount of data required to build a classifier increases exponentially with the number of features. 
\end{itemize}

\begin{figure}[h]
\centering
\includegraphics[width=0.9\textwidth]{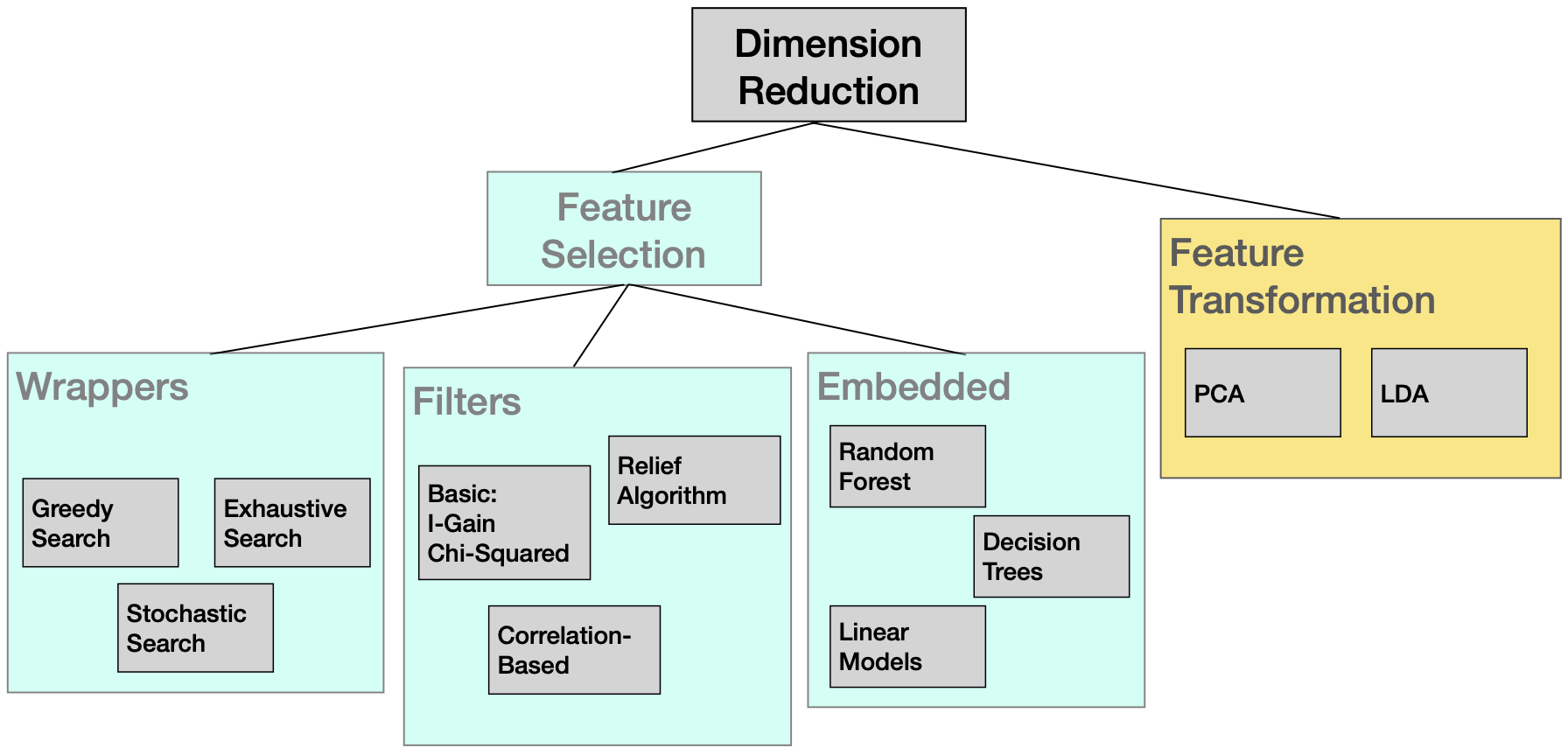}
\caption{An overview of the feature selection methods covered in this paper. For completeness, Principle Component Analysis and Linear Discriminant Analysis (which are not really feature selection methods) are also covered.}
\label{fig:Overview}
\end{figure}

Feature selection is most effective in the context of supervised machine learning (classification/regression) where the availability of labelled examples can drive the selection process. 
The methods we cover are summarised in Figure \ref{fig:Overview}. Other surveys of feature selection \cite{molina2002feature,guyon2003introduction} divide  feature selection methods into three categories and we follow the same structure:
\begin{itemize}
    \item \tb{Wrappers} are feature selection methods where the classifier is \emph{wrapped} in the feature selection process. This wrapping allows classification performance to drive the feature selection process. This has the advantage of tying the feature selection to classifier performance but this comes with a significant computational cost as very many classifier variants will be evaluated during the selection.
    \item \tb{Filters} cover methods that use criteria other than classifier performance to guide feature selection. Typically a filter provides a feature ranking and then a selection policy uses this ranking to select a feature subset.  
    \item \tb{Embedded} methods refer to any method where the feature selection \emph{emerges} as a by-product of the classifier training process. For instance, training a decision tree will almost always select a subset of the available features to build a tree. 
\end{itemize}

For completeness we also cover some dimensionality reduction methods that transform the data into a reduced dimension space rather than select a subset of features. In  section \ref{sec:FeatTrans} we cover Principle Component Analysis (PCA) and Linear Discriminant Analysis (LDA) for projecting data into lower dimension spaces. These are not feature selection methods in the sense that the original feature representation is left behind. 

Before working through the details of the Feature Selection methods the paper begins with a short discussion on the implications of The Curse of Dimensionality. Then the evaluation methodology used in the paper is presented in Section \ref{sec:method}.
The main material on Feature Selection methods is covered under the categories of Wrapper methods (Section \ref{sec:wrapper}), Filter strategies (Section \ref{sec:filters}) and Embedded methods (\ref{sec:embedded}). Section \ref{sec:FeatTrans} provides a brief description of PCA and LDA as already mentioned. 

\section{The Curse of Dimensionality} Before proceeding it is worth briefly discussing the implications of the Curse of Dimensionality for ML. The term was coined by Richard Bellman in 1961 and refers to a number of phenomena associated with data described by many features \cite{bellman2015adaptive}. For our purposes there are two theoretical issues. The first is that the number of samples required to build a model may increase exponentially with the number of features (dimensions). The second is that the variation in distance between arbitrary points \emph{decreases} as more dimensions are added. 

\begin{figure}[h]
\centering
\includegraphics[width=0.7\textwidth]{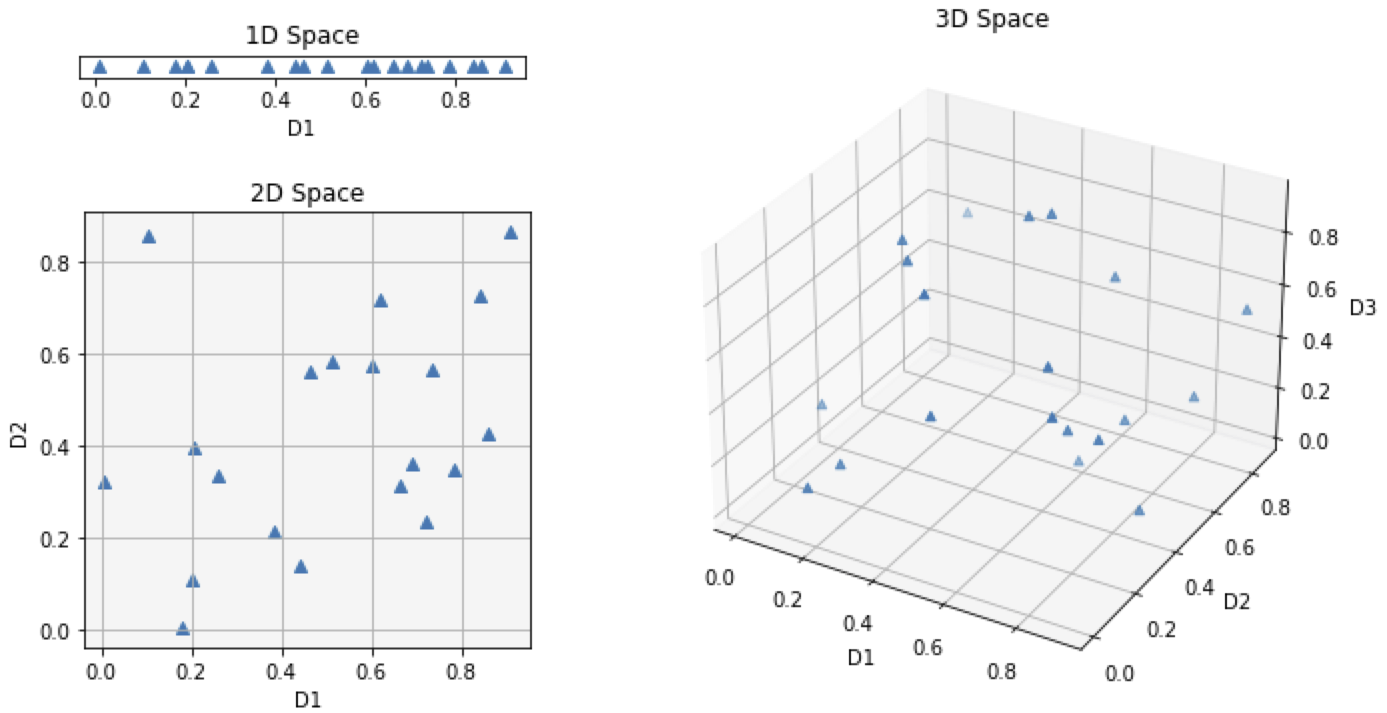}
\caption{20 random data points in 1D, 2D and 3D space. These plots show how data becomes increasingly sparse as the dimension is increased.}
\label{fig:1D2D3D}
\end{figure}

The first of these issues is illustrated in Figure \ref{fig:1D2D3D}. This figure shows plots of 20 random data points in 1D, 2D and 3D. As the dimension increases the data gets more and more sparse. A consequence of this is that the number of samples required to cover a phenomenon increases exponentially with dimension. This is less of a problem in practice because real data is unlikely to occupy all of the space, instead it will occupy a lower dimension manifold in the space. So the \emph{intrinsic} dimension of the data is likely to be considerably less than the full dimension \cite{cunningham2020knearest}. The PCA example in Figure \ref{fig:HarryPCA} also illustrates how the intrinsic dimension can be less than the number of features. 

\begin{figure}[h]
\centering
\includegraphics[width=0.5\textwidth]{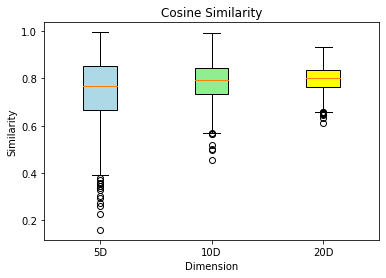}
\caption{A demonstration of how the variation in Cosine similarity reduces as the dimension of the data increases.}
\label{fig:3NN}
\end{figure}

The second issue is that, somewhat paradoxically, the more features used to describe the data the more similar everything appears. The box plots in Figure \ref{fig:3NN} illustrate this. The plots show the distribution of Cosine similarities between a probe point and 1,000 random points in 5D, 10D and 20D spaces. As the dimension increases the variation in similarity/distance decreases. 

In addition to these theoretical problems with having more features than is absolutely necessary, there are also practical problems to be considered. For most ML algorithms, the computational cost will increase with the number of features. As will the cost of gathering the data in the first place. So if the number of features can be reduced then there are strong theoretical and practical reasons for doing so. 

\section{Methodology}\label{sec:method}
The datasets used in the examples in this paper are summarised in Table \ref{tab:Data}. The Segmentation data set is from the UCI repository\footnote{\url{https://archive.ics.uci.edu/}}, the Penguins dataset was constructed from a larger dataset available on GitHub\footnote{\url{https://github.com/allisonhorst/palmerpenguins}} and the Harry Potter dataset was constructed by the authors based on the popular Top Trumps game\footnote{\url{https://toptrumps.us/collections/harry-potter}} to illustrate PCA in operation. The Segmentation data comes from an image segmentation task on outdoor images. The ground truth segments the images into seven classes and the instances are represented by 19 features. This is a good dataset for feature selection as there is some redundancy in the data. The Penguins dataset is a three class dataset described by four features \cite{gorman2014ecological}. The features are physical parameters (e.g. bill and flipper dimensions) and are strongly correlated so there is some redundancy. 

\begin{table}[ht]
\caption{Data Sets: Summary details.}
\label{tab:Data}

\begin{center}
\begin{tabular}{ l|r|r|r } 
Name & Samples & Features & Classes \\
\hline
Segmentation & 2,310 & 19 & 7 \\
Penguins & 333 & 4 & 3 \\
Harry Potter &22 & 5 & - \\
\end{tabular}
\end{center}
\end{table}

\begin{figure}[h]
\centering
\includegraphics[width=0.5\textwidth]{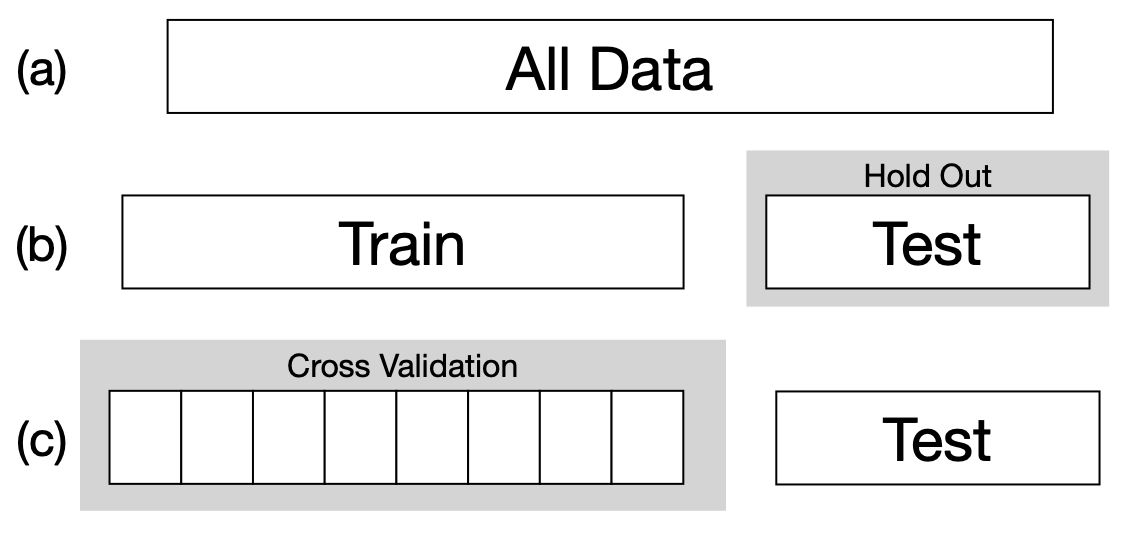}
\caption{Evaluation methodology. (a) If an estimation of generalisation accuracy is not required then all data can be used for all aspects of model development. (b) Test data can be held back from training to get an estimate of generalisation accuracy. (c) Cross validation can be used within the training data for Feature Selection.}
\label{fig:method}
\end{figure}

If we wish to get an assessment of generalisation performance for an ML system that might be deployed then we need to hold back some data for testing (option (b) in Figure \ref{fig:method}). If we wish to assess a few different feature selection alternatives as part of the model development then these should be tested within the confines of training data and cross validation is the most effective way to do this (option (c)). Most of the testing reported in this paper follows this pattern. Indeed, the evaluation testing may involve two levels of cross validation -- this strategy is not used here. It should be remembered that if the objective is to perform feature selection as part of the deployment of an ML system then all the available data can be used for feature selection (option (a) in Figure \ref{fig:method}). 

Unless otherwise stated, the classifier used in the evaluations in $k$-Nearest Neighbour ($k$-NN) \cite{cunningham2020knearest}. $k$-NN is used because it is probably the classifier in popular use most susceptible to noisy or redundant features. So the impact of feature selection will be most evident when it is used in evaluations. 

Before proceeding we need to introduce the notation that will be used throughout the paper. Assume we have a dataset $D$ made up of $n$ data samples. $D = \left<\mathbf{X},\mathbf{y}\right>$ where $\mb{y}$ are the class labels.
The examples are described by a set of features $F$ where $p=|F|$ so there are $n$ objects described by $p$ features. So the dimensions are $\mathbf{X}_{n \times p}$ and $\mathbf{y}_{p}$. 
The objective is to identify a subset $S \subset F$ that captures the important information in the dataset. In supervised ML the classifiers would work with data represented by a reduced set of features $\left<\mb{X}^\prime_{n \times k},\mb{y}\right>$ where $k = |S|$.

\section{Feature Selection using Wrappers} \label{sec:wrapper}

So the objective with feature selection is to identify a feature subset $S \subset F$ to represent the data. 
 If $|F|$ is small we could in theory try out all possible subsets of features and select the best subset. In this case \emph{`try out'} would mean training and testing a classifier using the feature subset. This would follow the protocol presented in Figure \ref{fig:method} (c) where cross-validation on the training data would identify a good feature subset and then this could be tested on the test data. 
However the number of possibilities is $2^p$ so exhaustive search quickly becomes impossible. 

\begin{figure}[t]
\centering
\includegraphics[width=0.6\textwidth]{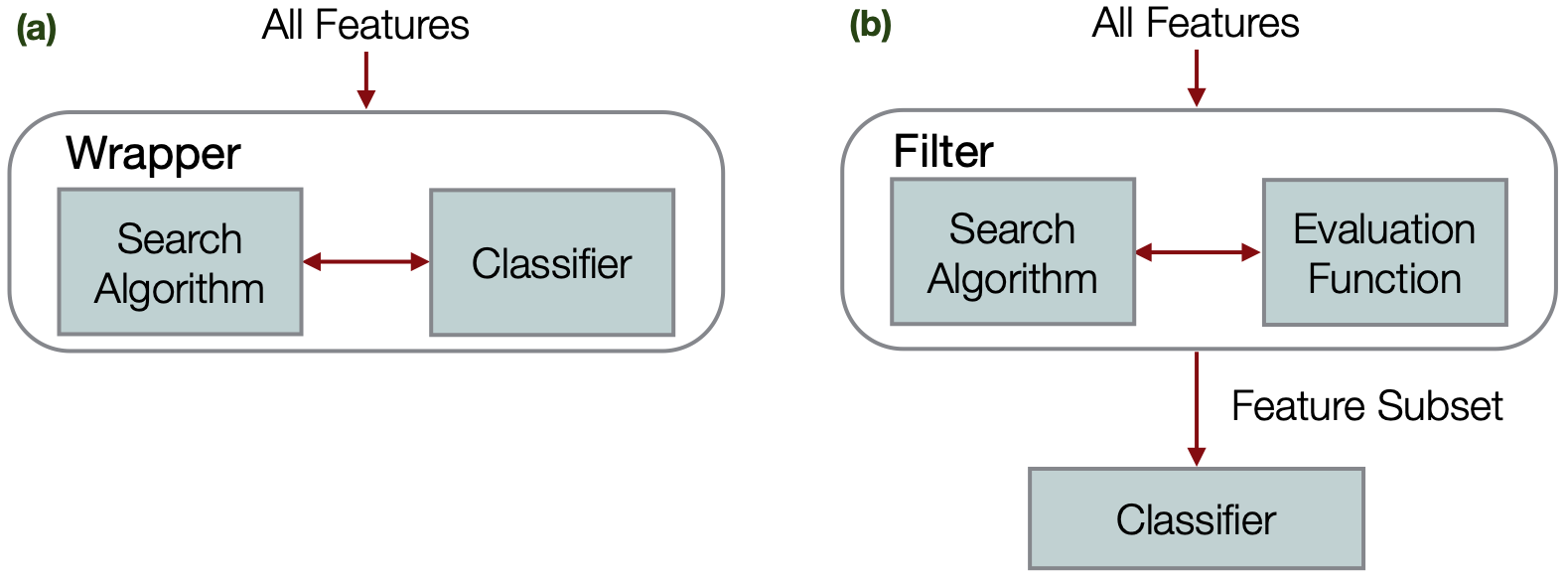}
\caption{Wrappers versus Filters: (a) With Wrappers the classifier is \emph{wrapped} in the search process. (b) A Filter  strategy uses a separate evaluation to score features}
\label{fig:WrapperFilter}
\end{figure}

\begin{figure}[h]
\centering
\includegraphics[width=0.7\textwidth]{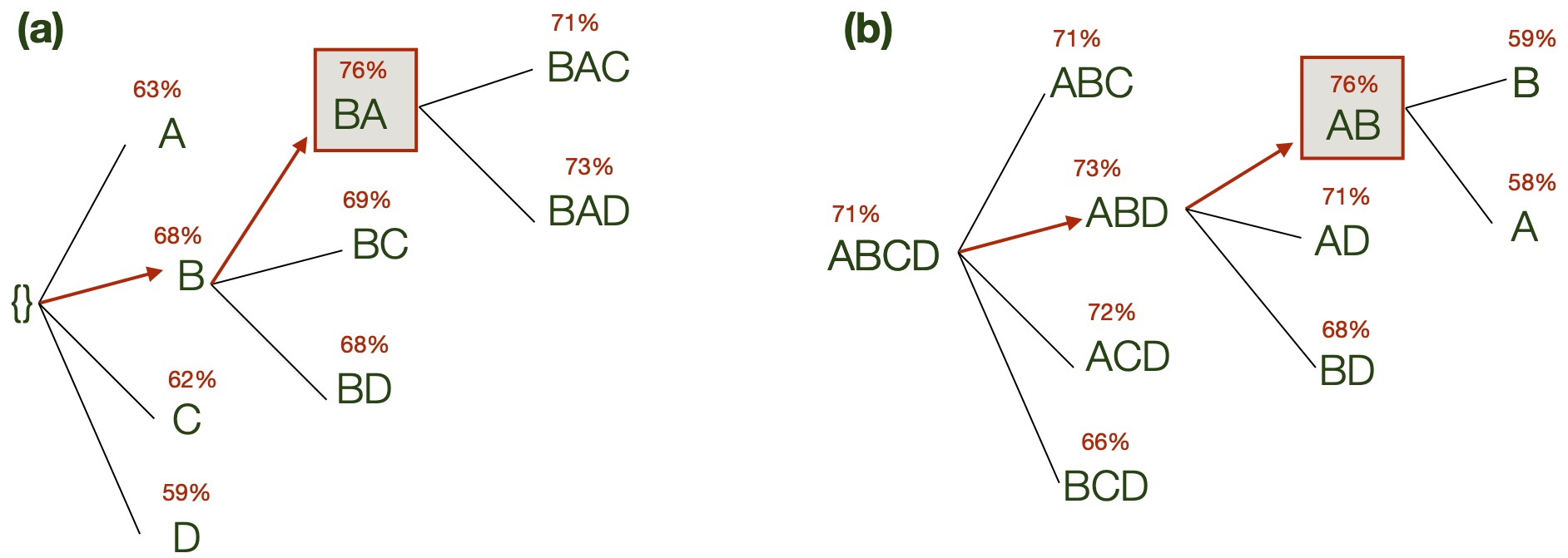}
\caption{Examples of feature subset selection using wrappers:  (a) Sequential Forward Selection (b) Backward Elimination.}
\label{fig:WrapperEx}
\end{figure}

Nevertheless this is how a Wrapper feature selection strategy works with the important modification that the search can be greedy or stochastic rather than exhaustive. The general idea is shown in Figure \ref{fig:WrapperFilter}(a), the classifier is \emph{wrapped} in the feature selection process, i.e. classifiers trained using the feature subsets are used in the search process. The feature subsets will be evaluated using hold-out testing or cross-validation testing on classifiers built using the data. The main search strategies used with Wrappers are:
\begin{itemize}
\item \tb{Exhaustive Search} evaluates every possible feature subset. If the number of features to be considered is small it will be possible to consider all feature combinations. However, if $p > 20$ there will be millions of feature subsets to considered and an exhaustive search will not be practical. 
    \item \tb{Sequential Forward Selection (SFS)}  starts with no features selected and all classifiers incorporating a single feature are considered (see Figure \ref{fig:WrapperEx} (a)). The best of these is selected and then two feature combinations including this feature are evaluated. This process proceeds, adding the winning feature at each step, until no further improvements can be made. 

\item \tb{Backward Elimination (BE)} proceeds in the opposite direction to FSS, it starts with all features selected, considers the options with one feature deleted, selects the best of these and continues to eliminate features. Again, the process is terminated when no improvements can be made. 

\item \tb{Stochastic Search} methods such as genetic algorithms or simulated annealing can readily be applied to Wrapper feature selection. Each state can be defined by a feature mask on which crossover and mutation can be performed \cite{loughrey2004overfitting}. Given this convenient representation, the use of a stochastic search for feature selection is quite straightforward although the evaluation of the fitness function (classifier accuracy as measured by cross-validation) is expensive.
\end{itemize}

Our exploration of Wrappers will focus on SFS and BE. These are greedy strategies that explore the search space of possible feature subsets as shown in Figure \ref{fig:WrapperEx}. SFS starts with an empty set and proceeds forward considering classifiers built on single features. The best of these is selected and then pairs of features incorporating this feature are considered. The process could terminate when the addition of a new feature doesn't result in any improvement. As the name suggests, Backward Elimination works in the opposite direction. It starts with a full set of features (Figure \ref{fig:WrapperEx} (b)) and eliminates the least useful feature at each step. For both SFS and BE, the feature subsets are evaluated using cross-validation on the training data. As stated in Section \ref{sec:method}, the classifier used is $k$-NN.

Both methods have their own advantages and disadvantages. SFS is inclined to require less computation as the models being evaluated are smaller, typically a classifier with a small number of features will take less time to train and test. SFS is inclined to select less features; this parsinomy is typically an advantage. On the other hand, because BE starts with larger feature sets, it can do a better job of assessing how features work in combination. 

\begin{figure}
     \centering
      \begin{subfigure}[b]{0.45\textwidth}
         \centering
         \includegraphics[width=\textwidth]{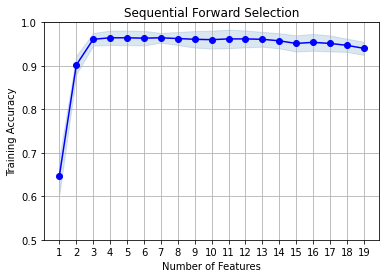}
       \caption{}
       \end{subfigure}
     \hfill
     \begin{subfigure}[b]{0.52\textwidth}
         \centering
         \includegraphics[width=\textwidth]{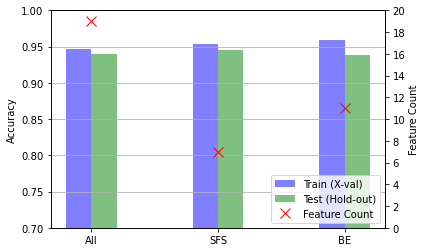}
         \caption{}
     \end{subfigure}
        \caption{Feature selection example using wrappers. (a) Accuracy on the training data as Sequential Forward Selection proceeds measured using cross-validation. (b) Accuracy estimates for feature subsets selected by SFS and BE. SFS selects 7 features and BE selects 11.}
        \label{fig:WrapperPython}
\end{figure}

In the Appendix a link is provided to code to run SFS and BE in Python. The evaluation is on the Segmentation dataset and the results are shown in Figure \ref{fig:WrapperPython}. On the left we see a plot of accuracy on the training set as the SFS proceeds. In this case the search is allowed to run to the end but it is evident that accuracy stops improving after seven features. 

The overall results for SFS and BE are shown in Figure \ref{fig:WrapperPython} (b). SFS selects seven features and 11 are selected by BE. Both feature subsets result in improved accuracy on the training data but only the SFS subset results in better accuracy on the test data. Indeed the gap between train and test accuracy for BE is evidence of overfitting -- the selection process has fitted too closely to the characteristics of the training data at the cost of generalisation accuracy. Indeed overfitting is recognised to be a problem with Wrapper-based feature selection \cite{loughrey2004overfitting}.

\section{Filter Strategies}\label{sec:filters}
Figure \ref{fig:WrapperFilter} (a) shows how Wrapper strategies use the classification algorithm in the feature selection process. Figure \ref{fig:WrapperFilter} (b) shows that Filter strategies do not use the classifier for feature selection, instead a separate evaluation function is used. The fact that Filters are independent of the classifier is a mixed blessing. It means that Filters can be much faster than Wrappers but the selected features may not be in tune with the inductive bias of the classifier. 

In the next subsection we provide some detail on the operation of a basic Filter strategy. Then we cover the Relief Algorithm and Correlation-Based Feature Selection, two Filter strategies that have received a lot of attention in recent years. 

\subsection{Basic Filters}\label{sec:basicF}
A basic Filter will entail a feature scoring mechanism and then a selection strategy based on these scores. The scoring mechanism needs to quantify how much information the feature has about the outcome. The selection strategy might be:
\begin{itemize}
\item Select the top ranked $k$ features,
\item Select top 50\%,
\item Select features with scores $> 50\%$ of the maximum score,
\item Select features with non-zero  scores.
\end{itemize}

\begin{figure}[h]
\centering
\includegraphics[width=0.6\textwidth]{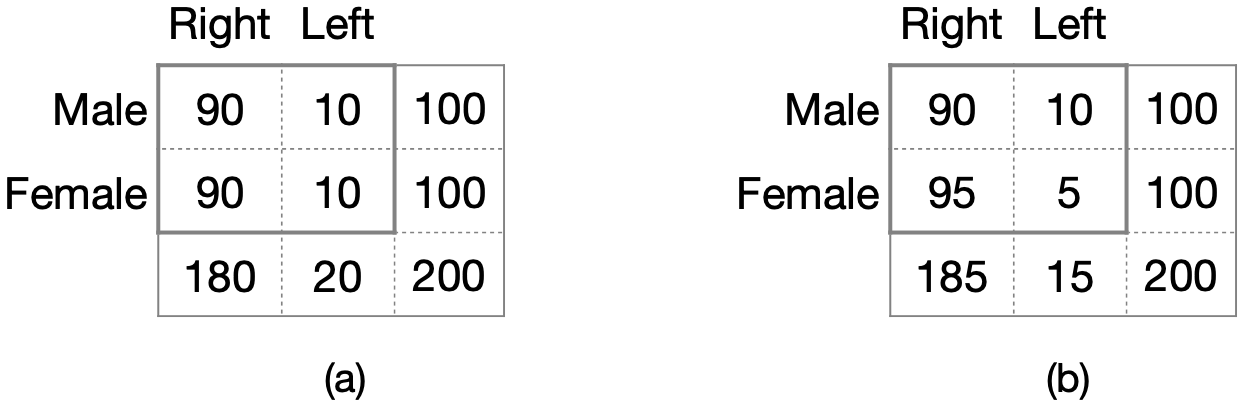}
\caption{Two contingency tables showing relationships between handedness and gender. If handedness is the class then in (a) it is independent of the gender feature, in (b) there is a dependence.}
\label{fig:HandednessEx}
\end{figure}

In this analysis we consider the Chi-square statistic and information gain for scoring. The Chi-square statistic is a measure of independence between a feature and the class label. If samples are organised into a contingency table as shown in Figure \ref{fig:HandednessEx}, how different are the cell counts to what would be observed by chance? The data in Figure \ref{fig:HandednessEx} (a) suggests that handedness is independent of gender because the proportions are the same. The data in (b) suggests that gender is predictive of handedness. The Chi-square statistic allows us to quantify this:

\begin{equation}
    \chi^2 = \sum_{i=1}^m
    \frac{(O_i - E_i)^2}{E_i}
\end{equation}

The statistic is a sum over the $m$ cells. For each cell we consider the difference between the observed count $O_i$ and the expected count $E_i$ if the feature and the class were independent. In Figure \ref{fig:HandednessEx} (a) this difference would be zero because the feature and the class are independent. In (b) there would be a difference so the statistic would be positive. In general, the greater the dependence the larger the statistic. If the feature values are numeric rather than categorical then the feature values can be binned to enable the construction of the contingency table \cite{jin2006chi-sq}. 

Information gain is an alternative information-theoretic measure quantifying the information a feature contains about a class 
\cite{kelleher2020fundamentals}. In Figure \ref{fig:HandednessEx} (b) by knowing the gender we \emph{gain} information about handedness. In a binary classification scenario, let's assume the probability of a positive and negative outcomes are respectively $p$ and $q$. Then the entropy of a dataset based on these proportions is:
\begin{equation}
    H(D) = -p\log_2(p) - q\log_2(q)
\end{equation}

then the information gain for any feature $f$ in the dataset in terms of the class label is:

\begin{equation}
    IG(D,f) = H(D) - \sum_{v \in values(f)} \frac{|S_v|}{S}H(D_v)
\end{equation}

\begin{figure}[ht]
     \centering
      \begin{subfigure}[c]{0.49\textwidth}
         \centering
         \includegraphics[width=\textwidth]{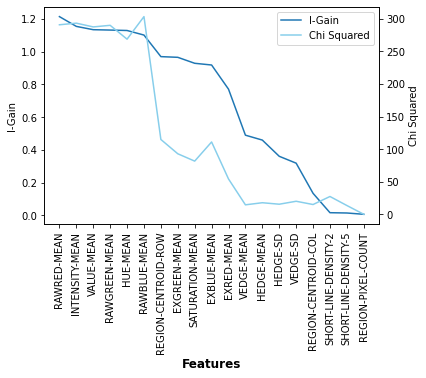}
         \caption{}
       \end{subfigure}
     \hfill
     \begin{subfigure}[c]{0.49\textwidth}
         \centering
         \includegraphics[width=\textwidth]{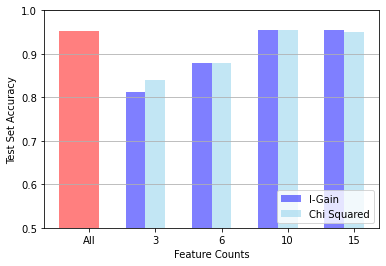}
         \caption{}
     \end{subfigure}
        \caption{Feature selection using filters. (a) I-Gain and Chi-square scores for the 19 features in the segmentation dataset. (b) Accuracy estimates for top-$n$ features.}
        \label{fig:Filter-ex}
\end{figure}

As with the Chi-square statistic, information gain (I-Gain) allows us to rank features for the purpose of feature selection. This is illustrated in Figure \ref{fig:Filter-ex} (a). This shows the Segmentation features ranked by both measures. A link for this code is provided in the Appendix. The plot shows the scores sorted by I-Gain score. It is clear that the scores are well correlated (Pearson correlation score of 0.86) so feature subsets selected based on these scores should be reasonably similar. 

This does prove to be the case when we look at the performance of classifiers built with feature subsets based on these rankings. In  Figure \ref{fig:Filter-ex} (b) we see the results of a range of top $k$ selection policies ($k = 3,6,10,15$). At $k = 10$ both scores select a feature subset that produces accuracy on the test set equivalent to that obtained with the full feature set. The evaluation strategy here conforms to pattern (b) in Figure \ref{fig:method}, the feature scoring is done using the training data and then tested on the test set. 

\subsubsection{Combining Filters and Wrappers}\label{sec:hybrid}
The results in Figure \ref{fig:Filter-ex} show that initially performance improves as features are added based on the ranking generated by the filters. However, this improvement tails off and eventually no improvement results from the addition of `poorer' features. Indeed these features may damage performance. This suggests a hybrid Filter/Wrapper strategy whereby a Filter is used to rank the features and then  a Wrapper is used to identify the optimum feature subset. 
 
\begin{figure}[htb]
\centering
\includegraphics[width=0.65\textwidth]{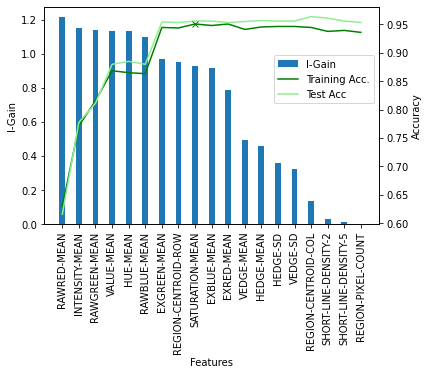}
\caption{A hybrid Wrapper-Filter strategy. Features are ranked using I-Gain and then the performance of subsets based on this ranking is evaluated. The training set accuracy of  the best performing feature subset is marked with an \textsf{X}. }
\label{fig:IG-Wrap}
\end{figure}

This hybrid strategy is shown in operation in Figure \ref{fig:IG-Wrap}. The dataset has been split into train and test sets of equal size. The training set is used to estimate I-Gain scores for all features, these are shown in blue. Then classifiers are trained with feature sets of increasing size. The performance of these feature subsets are scored on the training set using cross validation and on the test set using hold-out testing. After the addition of nine features the performance on the training set stops improving (indicated with a green \textsf{X}). This hybrid strategy would select this as the optimum feature subset. 

\subsection{Relief Algorithm and Variants} 
The Relief family of algorithms deserve mention because, as Filter methods, they have the advantage of speed while scoring features in the context of other features \cite{kira1992feature,kononenko1994estimating}. Relief algorithms belong to the $k$-NN paradigm in ML. A $k$-NN analysis is used to score or weight features. The idea is to take each sample in the dataset (or a subset) and find its nearest neighbour of the same class and nearest unlike neighbour - these are termed the \textsf{nearHit} and the \textsf{nearMiss}. Then the general principle is:
\begin{itemize}

\item For the \textsf{nearMiss} ($nM$) increment the feature weights; weights for unmatching features will be incremented more. 

    \item For the \textsf{nearHit} ($nH$) decrement the feature weights; weights for unmatching features will be decremented more. 
\end{itemize}

\begin{figure}[h]
\centering
\includegraphics[width=0.45\textwidth]{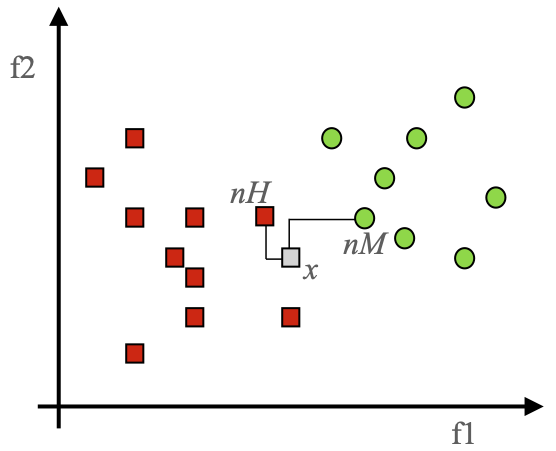}
\caption{An illustration of the principles underlying Relief in a 2D space. $x$ is the query, $nM$ is the is the \textsf{nearMiss}, $nH$ is the \textsf{nearHit}. f1 should be preferred over f2. }
\label{fig:Relief}
\end{figure}

The idea is that this will pull matching instances closer together and push unmatching instances apart. 
This is illustrated in the 2D example shown in Figure \ref{fig:Relief}. The weights should be adjusted to bring $x$ and $nH$ closer together while pushing $nM$ away. Comparing $x$ and $nM$, they match more on f2 than on f1 so the weight for f1 will be incremented more. Considering $x$ and $nH$ the opposite happens, f2 is decremented more than f1, so the overall effect is that f1 scores better than f2.
This is achieved with the following update function where $x$ is the query, $nM$ is the \textsf{nearMiss}, $nH$ is the \textsf{nearHit} and $w_f$ is the weight for feature $f$. 
\begin{equation}
    w_f \leftarrow w_f -(x_f - nH_f)^2+(x_f - nM_f)^2
\end{equation}
This achieves what we want because the feature value differences for matching features will be smaller than those for unmatching features. 

This process is repeated multiple times to produce a set of feature weights with `good' features having high weights. Implementation details of the Relief family of algorithms are provided by Urbanowicz \emph{et al.} \cite{urbanowicz2018benchmarking}.
A link to sample code for running Relief is provided in the Appendix. The results are summarised in Figure \ref{fig:Relief-eval}.  On the left we see Relief scores for the segmentation dataset compared with I-Gain scores. The Spearman correlation is 0.91 so both scores are well correlated as is evident from the plot. The plot suggests that there is a clear partition between the first 11 features and the remainder so we test the accuracy of a classifier built using just these 11 features. The results on the right show that this does result in a small improvement in accuracy compared with using all features. 

\begin{figure}[h]
     \centering
      \begin{subfigure}[c]{0.6\textwidth}
         \centering
         \includegraphics[width=\textwidth]{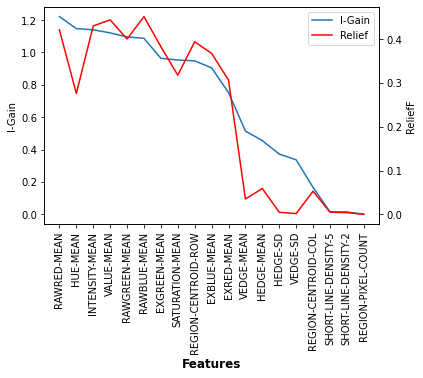}
       \end{subfigure}
     \hfill
     \begin{subfigure}[c]{0.3\textwidth}
         \centering
         \includegraphics[width=\textwidth]{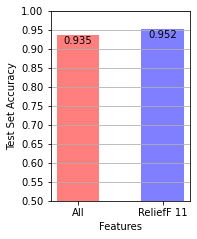}
     \end{subfigure}
        \caption{Feature selection using ReliefF. (a) I-Gain and ReliefF scores for the 19 features in the segment dataset. (a) Accuracy estimates using All features and the top 11 as selected using ReliefF.}
        \label{fig:Relief-eval}
\end{figure}

\subsection{Correlation-Based Feature Selection}
Correlation Based feature selection (CFS) is a filter strategy that relies on the principle that "A good feature subset is one that contains features highly correlated with (predictive of) the class, yet uncorrelated with (not predictive of) each other" \cite{hall1999correlation}. The feature-class correlation indicates how representative of the class that feature is while the feature-feature correlation indicates any redundancies between the features. CFS works by assigning a merit value based on feature-class and feature-feature correlations to each feature subset which becomes the measure by which subsets are evaluated.
\begin{figure}[h]
\centering
\includegraphics[keepaspectratio, trim = 20 0 20 0, clip, width=\textwidth]{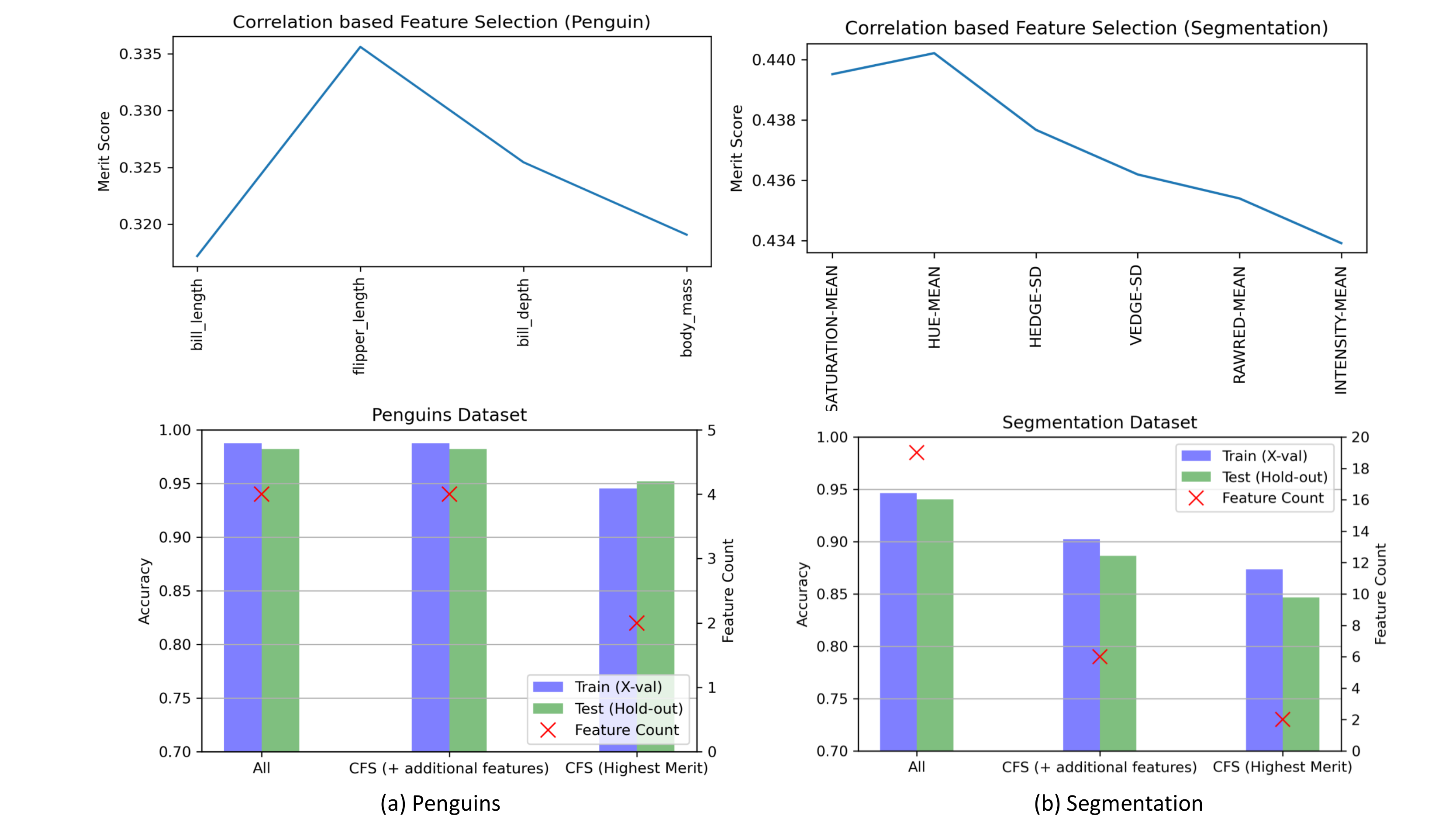}
\caption{Feature Selection using CFS. The graphs on top show how the merit scores change as the selected features are added to the selected set. The bar charts show accuracy scores for the different feature subsets. In both cases the subset with the highest merit score does not have the highest accuracy.} \label{fig:CFS}
\end{figure}
The merit score for a feature subset $S$ is: 
\begin{equation}
    M_{S} = \frac{k\overline{r_{cf}}}
    {\sqrt{k+k(k-1)\overline{r_{ff}}}}
    \label{eqn:MSTS}
\end{equation}
where $k = |S|$, $\overline{r_{ff}}$ is the average correlation between the features values in the subset ($f \in S)$ and
 $\overline{r_{cf}}$ is the average correlation between the  values for the selected features and the class label. 
 
The correlations can be measured using techniques such as symmetrical uncertainty based on information gain, feature weighting based on the Gini-index, or by using the minimum description length (MDL) principle \cite{hall1999correlation}. Information gain based methods work well and hence the symmetrical uncertainty score is commonly used in CFS implementations.

CFS can work with any search strategy in a similar way to wrapper strategies, but rather than evaluating based on accuracy as in Section \ref{sec:wrapper}, the evaluation will be based on the merit score. For example, when using sequential forward selection, all single features merit will be evaluated from which the best will be selected. All two feature combinations which include this first feature will then be evaluated using the merit score and so on until there is no more improvement in the merit score.

CFS shows a tendency to favour small subsets with moderate accuracy. The CFS implementation provided by \cite{li2018feature} utilises a Best First search which continues the search until five consecutive non-improving feature subsets are found. Therefore, even after the merit starts decreasing, the search continues. Figure \ref{fig:CFS} shows the benefit of this as it can be seen that for both the segmentation and the penguin datasets, the hold-out accuracy from using the extra features improves in comparison to evaluating the performance at the point where merit score is highest. In both datasets, CFS has led to a drop in the hold-out accuracy compared to using the full feature subset, however has selected a smaller subset than other techniques such as Relief-F. Hence CFS may be a good technique for feature reduction. A link to sample code for running CFS is provided in the Appendix.

\section{Embedded Methods}\label{sec:embedded}
In this section we cover feature selection methods that \emph{emerge} naturally from the classification algorithm or arise as a side effect of the algorithm. We will see that with Decision Trees and Logistic Regression feature selection can be an integrated part of the model building process. Then with Random Forest, we will see how feature importance scores can easily be generated from the model. 

\subsection{Decision Trees} 
\begin{figure}[h]
\centering
\includegraphics[width=0.8\textwidth]{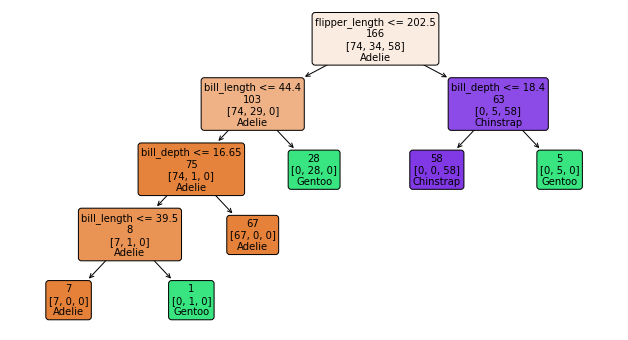}
\caption{A decision tree for the Penguins dataset. While the data is described by four features only three are selected.}
\label{fig:tree}
\end{figure}

The construction of a Decision Tree from a data set will very often entail feature selection as some of the features will not appear in the tree. Features not included in the tree are effectively selected out. We show an example of this on the Penguins dataset in Figure \ref{fig:tree}.

In this example the dataset has been divided 50:50 into train and test sets. This tree has been trained on the training data and has 93\% accuracy on the test data (see links in Appendix). This dataset has four features, \textsf{flipper\_length, bill\_length, bill\_depth} and \textsf{body\_mass}. It is clear from the tree in Figure \ref{fig:tree} that three of the four features are selected, \textsf{body\_mass} is not selected. 

This tree has been constructed with the default \textsf{scikit-learn} parameters so there is no pruning. It is normal in Decision Tree learning to constrain (i.e. prune) the size of the tree to prevent overfitting. The use of pruning to prevent overfitting will push the feature selection further as even less features will be selected in smaller trees. 

\subsection{ Logistic Regression: Lasso}\label{sec:reg}
In multivariate linear models such as linear regression or logistic regression, feature selection can be achieved as a side effect of regularization. In ML regularization refers to mechanisms designed to simplify models in order to prevent overfitting. Thus regularization can cause features to be deselected. Elastic net and Lasso are popular regularization methods for linear models. Here we will provide an overview of how Lasso works \cite{tibshirani1996regression} and present examples of Lasso in operation.  Starting with the basics, a multivariate regression model works as follows:
\begin{equation}
\begin{split}
    y =f(\mathbf{x}) & = \beta_0 + \sum_{i=1}^p \beta_i x_i \\
    & = \beta_0 + \mathbf{\beta x}
\end{split}
\end{equation}
The dependent variable $y$ is a linear function of the input features; for each feature $x_i$ the weight of that feature is determined by the corresponding $\beta_i$ parameter. For binary classification problems ([0,1] labels) we can use logistic regression where the dependent variable is the log odds that an outcome variable is 1. If $pr$ is the probability that the label is 1 then $\frac{pr}{1-pr}$ is the odds.
\begin{equation}
ln\left(\frac{pr}{1-pr} \right) = \beta_0 + \mathbf{\beta x}
\end{equation}

So logistic  regression provides a class probability:
\begin{equation}\label{eqn:logistic}
    pr  = \frac{1}{1 - e^{(\beta_0 + \mathbf{\beta x})}}
\end{equation}

Regularization prevents overfitting by limiting model capacity; this is done by limiting the size of weights. The two options are L$_1$ or L$_2$ regularization:
\begin{equation}\label{eqn:Lasso}
   \text{L}_1: \quad 
\sum_{i=1}^p |\beta_i|  < t
\end{equation}

\begin{equation}\label{eqn:L2_reg}
   \text{L}_2: \quad 
\sum_{i=1}^p \beta_i^2  < t
\end{equation}

So the $\beta$ parameters in (\ref{eqn:logistic}) are fitted to the training data subject to the constraints in (\ref{eqn:Lasso}) or in (\ref{eqn:L2_reg}). It transpires that when an L$_1$ regularization is used the weaker weights will go to zero, i.e. those features will be deselected. There is an excellent explanation of \emph{why} this happens in the original Lasso paper by Tibshirani \cite{tibshirani1996regression}.

\begin{figure}
     \centering
      \begin{subfigure}[c]{0.3\textwidth}
         \centering
         \includegraphics[width=\textwidth]{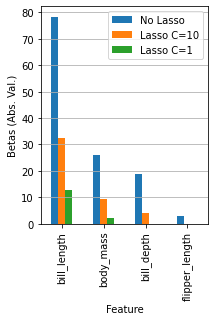}
         \caption{Penguins}
       \end{subfigure}
     \hfill
     \begin{subfigure}[c]{0.6\textwidth}
         \centering
         \includegraphics[width=\textwidth]{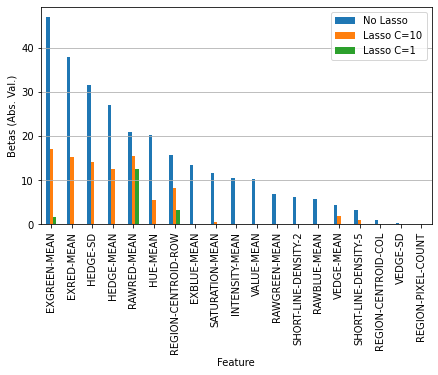}
         \caption{Segmentation}
     \end{subfigure}
        \caption{The impact of Lasso on the Segmentation and Penguins datasets. Lasso reduces the magnitude of the $\beta$ parameters; some parameters get reduced to zero. Results for Lasso with the default regularization (C=1) and milder regularization (C=10) are shown.}
        \label{fig:Lasso-feat}
\end{figure}
\begin{figure}
     \centering
      \begin{subfigure}[c]{0.49\textwidth}
         \centering
         \includegraphics[width=\textwidth]{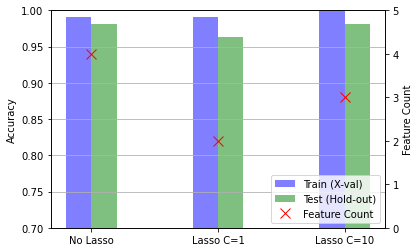}
         \caption{Penguins}
       \end{subfigure}
     \hfill
     \begin{subfigure}[c]{0.49\textwidth}
         \centering
         \includegraphics[width=\textwidth]{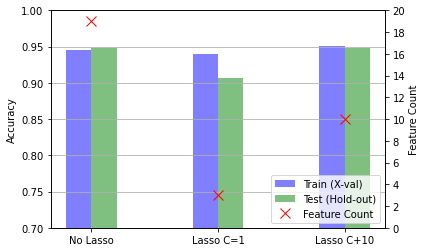}
         \caption{Segmentation}
     \end{subfigure}
        \caption{The impact of Lasso on train and test accuracy. Results for Lasso with the default regularization (C=1) and milder regularization (C=10) are shown.}
        \label{fig:Lasso-acc}
\end{figure}

To demonstrate this on our sample datasets we reduce them to binary classification problems to make the overall process  more transparent. However, feature selection using Lasso also works with multiclass problems. The results are shown in Figures \ref{fig:Lasso-feat} and \ref{fig:Lasso-acc}. Because the datasets have been reduced to just two classes (Cement and Window for Segmentation and Adelie and Chinstrap for Penguins) the accuracies are higher than for the multi-class scenario. The extent of the feature reduction with Lasso is controlled by the regularization parameter C. Results are included for two levels of regularization, C=10 and C=1. C=10 results in less regularization so more features are retained. In both cases the default regularization results in too much feature reduction and generalization accuracy is reduced. For the Penguins dataset just two features are retained while three are retained in the Segmentation dataset (see Figure \ref{fig:Lasso-feat}).
The milder regularization retains more features resulting in no loss of generalization accuracy.

\subsection{Random Forest Feature Importance}
Ensembles, the idea that a  committee of classifiers will be more accurate than a single classifier, are central to modern ML. It has been known for hundreds of years that a committee of decision makers will be better than  an individual
\cite{de1785essai}. In ML an ensemble of classifiers voting on an outcome will be more accurate than a single classifier provided some conditions are met. Two ensemble methods, gradient boosting \cite{friedman2002stochastic} and random forests  \cite{breiman2001random} represent the state of the art in supervised ML. Breiman \cite{breiman2001random} has shown that, as a supplementary benefit of the effort required to build the random forest, the ensemble offers estimates of generalisation accuracy and feature importance. In order to explain how these estimates of feature importance work we need to go into some detail on how random forests work. 

In order for an ensemble to be effective, there needs to  be some diversity among the ensemble members. If not the ensemble will be no better than the individual members (classifiers). If we have a dataset $\mathbf{X}_{n \times p}$  of $n$ samples described by $p$ features  with class labels $\mathbf{y}_{p}$ then we have two basic strategies for training diverse classifiers:
\begin{itemize}
    \item \textbf{Bagging:} We can train each classifier with different training sets drawn at random with replacement from the available data. If these `bagged' training sets are of size $n$ then some samples will be selected multiple times and some  not at all. For each training set, roughly 37\% of samples will not be selected, i.e. out-of-bag (OOB). 
    \item \textbf{Random Subspacing:} The other strategy for ensuring diversity is to work with subsets of features rather than subsets of samples. With random forest, only a subset of features are available for consideration at each split point in the construction of a tree. This subset might be quite small, for instance $\sqrt{p}$.
\end{itemize}
Random forest (RF) uses both of these strategies together to ensure diversity. A typical RF could contain 1,000 trees. The samples that are OOB in these trees can be used to get an estimate of generalisation accuracy for the RF without the need to hold back test data from the training process. These OOB samples can be used to generate feature importance scores. For each of the $p$ features in the OOB samples for a given tree, the values for that feature can be randomly permuted and the revised classifications for those samples can be saved. Then, for the whole ensemble, the impact of this feature value shuffling can be assessed. 

If a feature is not important, then the impact of this shuffling will be minimal. For predictive features this shuffling will have a significant impact. Furthermore, and this is important for feature selection, this feature importance is being assessed in the context of other features. This contrasts with the basic filter methods described in section \ref{sec:basicF} where features are scored in isolation.  

\begin{figure}
     \centering
      \begin{subfigure}[c]{0.3\textwidth}
         \centering
         \includegraphics[width=\textwidth]{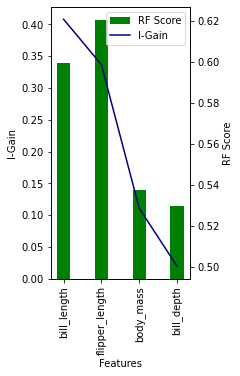}
        \caption{}

       \end{subfigure}
     \hfill
     \begin{subfigure}[c]{0.6\textwidth}
         \centering
         \includegraphics[width=\textwidth]{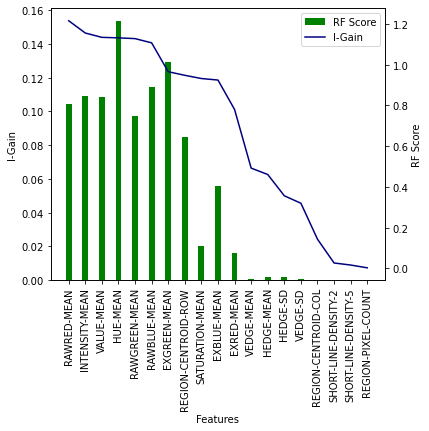}
         \caption{}
     \end{subfigure}
        \caption{Random forest feature importance. (a) RF feature importance scores for the Penguins dataset compared with I-Gain. (b) RF and I-Gain scores for the segmentation dataset.}
        \label{fig:RF-FI}
\end{figure}

RF feature importance scores for the Penguin and Segmentation datasets are shown in Figure \ref{fig:RF-FI}. I-Gain scores are also shown and these correlate well with the RF scores (0.8 for Penguins and 0.92 for Segmentation). For the Penguins dataset, the RF score ranks \textsf{flipper\_length} highest. Because RF feature importance considers features in context, this may indicate that this feature contains information not available in the other features. The same may be true for the \textsf{HUE\_MEAN} and \textsf{EXGREEN\_MEAN} features in the Segmentation data which are also given higher rankings.

\section{Feature Transformations} \label{sec:FeatTrans}
So far we have focused on feature ranking and feature subset selection methods as that is the main focus of the paper. However, it is also worth mentioning feature transformation methods such as Principal Component Analysis (PCA) as these methods are also used for dimension reduction. 

The dominant feature transformation technique is Principal Components Analysis (PCA) that transforms the data into a reduced space that captures most of the variance in the data. PCA is an unsupervised technique in that it does not take class labels into account. By contrast Linear Discriminant Analysis (LDA) seeks a transformation that maximises between-class separation (Section \ref{sec:LDA}).

As with feature selection we are concerned with datasets of $n$ objects described by $p$ features. Unfortunately, with feature transformation methods, it is not unusual to represent the data as a feature-object matrix $\mathbf{X}_{p\times n}$ where each column represents an object. This contrasts with the object-feature format $\mathbf{X}_{n\times p}$ popular in supervised ML where each row is an object. For consistency with the rest of this paper we will stick with the object-feature format $\mathbf{X}_{n\times p}$. The objective with Feature Transformation is to transform the data into another set of features $F^\prime$ where $k = |F^\prime|$ and $k < p$, i.e. $\mathbf{X}_{n\times p}$ is transformed to $\mathbf{X^\prime}_{n\times k}$. Typically this is a linear transformation $\mathbf{W}_{p\times k}$
that will transform each $\mathbf{x}_i$ to $\mathbf{x}^\prime_i$ in $k$ dimensions.

\begin{equation}\label{eqn:feat-trans}
\mathbf{x}^\prime_i = \mathbf{x}_i\mathbf{W}
\end{equation}
\subsection{Principal Component Analysis}\label{sec:PCA}

In PCA the transformation described in Equation \ref{eqn:feat-trans} is achieved so that feature $f_1^\prime$ is in the dimension in which the variance on the data is maximum, $f_2^\prime$ is in an orthogonal dimension where the remaining variance is maximum and so on.
Central to the whole PCA idea is the covariance matrix of the data 
\cite{hotelling1933analysis}. The diagonal terms in $\mathbf{C}$ capture the variance in the individual features and the off-diagonal terms quantify the covariance between the corresponding pairs of features. The objective with PCA is to transform the data so that the covariance terms are
zero, i.e. the new dimensions are independent. The overall process is as follows:



\begin{enumerate}
    \item Calculate the means of the columns of \textbf{X}.
    \item Subtract the column means from each row of \textbf{X} to create the \emph{centred matrix} \textbf{Z}.
    \item Calculate the covariance matrix $\mathbf{C} = \frac{1}{n -1} \mathbf{Z}^\mathsf{T}\mathbf{Z}$.
    \item Calculate the eigenvectors and eigenvalues of the covariance matrix $\mathbf{C}$. 
    \item Examine the eigenvalues in descending order to determine the number of dimensions $k$ to be retained - this is the number of principle components. 
    \item The top $k$ eigenvectors make up the columns of the transformation matrix $\mathbf{P}$ which has dimension $(p \times k)$.
    \item The data is transformed by $\mathbf{X}^\prime = \mathbf{Z}\mathbf{P}$ where $\mathbf{X}^\prime$ has dimension $(n \times k$).
\end{enumerate}

The $i^{th}$ diagonal entry in $\mathbf{C}$ quantifies the variance of the data in the direction of the corresponding principal component. Dimension reduction is achieved by discarding the lesser principal components, i.e. $\mathbf{P}$ has dimension $(p \times k)$ where $k$ is the number of principal components retained.

We illustrate PCA in operation with the example shown in Table \ref{tab:Harry} and Figure \ref{fig:HarryPCA}. The code for this is linked in the Appendix. This example is based on the Top Trump childrens' game. Each object is a card representing a Harry Potter character described by five features. In this dataset there are 22 cards so $n = 22$ and $p = 5$. Figure \ref{fig:HarryPCA}(a) shows the variance in the data captured by the first four principal components (PCs). The first two PCs together account for 81\% of the variance in the data so when in Figure \ref{fig:HarryPCA}(b) we plot the data in terms of these two PCs most of the variance in the data is retained. 

\begin{table}[h]
\caption{Sample Harry Potter data for use in PCA.}
\label{tab:Harry}

\begin{center}
\begin{tabular}{ l|r|r|r|r|r } 
Name & Magic & Cunning & Courage & Wisdom & Temper \\
\hline
Harry Potter & 62 & 21 & 42 & 26 & 7 \\
Hermione Granger & 60 & 16 & 40 & 73 & 2 \\
Ron Weasley & 45 & 14 & 40 & 22 & 4 \\
Prof. Dumbledore & 105 & 24 & 39 & 82 & 0 \\
Prof. Snape & 85 & 24 & 19 & 71 & 7 \\
...  & ... & ... & ... & ... & ...\\
\end{tabular}
\end{center}
\end{table}

\begin{figure}
     \centering
      \begin{subfigure}[c]{0.4\textwidth}
         \centering
         \includegraphics[width=\textwidth]{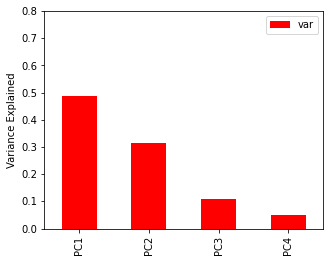}
        \caption{}

       \end{subfigure}
     \hfill
     \begin{subfigure}[c]{0.55\textwidth}
         \centering
         \includegraphics[width=\textwidth]{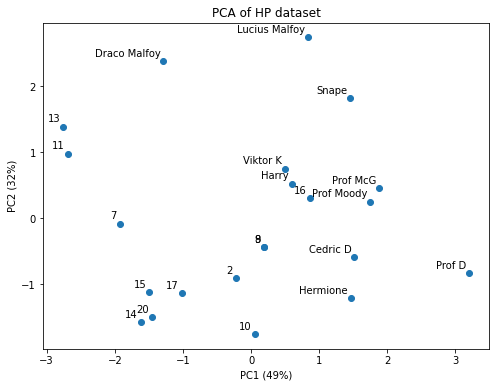}
         \caption{}
     \end{subfigure}
        \caption{PCA on the Harry Potter dataset shown in Table \ref{tab:Harry}. (a) The variance explained by the first four principal components (original data has five features). (b) A 2D plot of the data in the first two principal components.}
\label{fig:HarryPCA}
\end{figure}

While PCA is the established method for unsupervised dimension reduction there are other methods also in use. Singular Value Decomposition (SVD), a closely related matrix factorization method, is popular in text analytics 
\cite{cunningham2008dimension} and Latent Dirichlet Allocation
\cite{blei2003latent} is popular for topic modelling. 

\subsection{Linear Discriminant Analysis}\label{sec:LDA}

PCA on the Penguins dataset is shown in Figure \ref{fig:PCA-LDA}(a). Given that PCA is designed to project the data into dimensions that capture the variance in the data it should not be surprising that it does not do a great job of separating the classes.  Figure \ref{fig:PCA-LDA}(b) shows Linear Discriminant Analysis (LDA) on the same dataset. LDA takes the class labels into account and seeks a projection that maximises the separation between the classes. The objective is to uncover a transformation that will maximise between-class separation and minimise within-class separation. To do this we define two scatter matrices,  $\mathbf{S}_B$  for between-class separation and  $\mathbf{S}_W$  for within-class separation:

\begin{figure}
     \centering
      \begin{subfigure}[c]{0.4\textwidth}
         \centering
         \includegraphics[width=\textwidth]{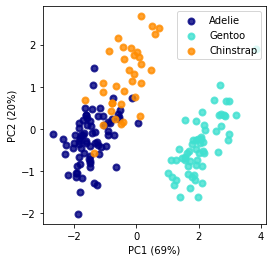}
        \caption{PCA}

       \end{subfigure}
     \hfill
     \begin{subfigure}[c]{0.4\textwidth}
         \centering
         \includegraphics[width=\textwidth]{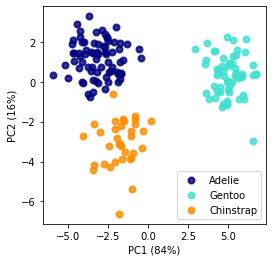}
         \caption{LDA}
     \end{subfigure}
        \caption{PCA and LDA on the Penguins dataset. PCA seeks to maximize the variance captured by the two components whereas LDA seeks to maximise the separation between the classes.}
        \label{fig:PCA-LDA}
\end{figure}

\begin{equation}
    \mathbf{S}_B = \sum_{c \in C} n_c(\mu_c - \mu) (\mu_c - \mu)^{\mathsf{T}}
\end{equation}

\begin{equation}
    \mathbf{S}_W = \sum_{c \in C} \sum_{j:y_j = c }n_c(x_j - \mu_c) (x_j - \mu_c)^{\mathsf{T}}
\end{equation}
where $n_c$ is the number of objects in class $c$, $\mu$ is the mean of all examples and $\mu_c$ is the mean of all examples in class $c$:

\begin{equation}
    \mu = \frac{1}{n}\sum_{i=1}^n x_i \qquad
       \mu_c = \frac{1}{n_c}\sum_{j:y_j = c} x_j
\end{equation}

The components within these summations $\mu,\mu_c,x_j$ are vectors of dimension $p$ so $\mathbf{S}_B$ and $\mathbf{S}_W$  are matrices of dimension $p\times p$.
The objectives of maximising between-class separation and minimising within-class separation can be combined into a single maximisation called the Fisher criterion \cite{fisher1936use,fukunaga2013introduction}:
\begin{equation}\label{eqn:LDA}
    \mathbf{W}_{LDA} = \arg\max_{\mathbf{W}}\frac{|\mathbf{W}^{\mathsf{T}}\mathbf{S}_B\mathbf{W}|}
    {|\mathbf{W}^{\mathsf{T}}\mathbf{S}_W\mathbf{W}|}
\end{equation}

i.e. find $\mathbf{W} \in \mathbb{R}_{p \times k}$
so that this fraction is maximised ($ |A| $ denotes the  determinant of matrix $A$). This matrix $\mathbf{W}_{LDA}$ provides the transformation described in equation
\ref{eqn:feat-trans}. While the choice of $k$ is again open to question it is sometimes selected to be $k = |C|-1$, i.e. one less than the number of classes in the data. Solving the optimization problem presented in equation \ref{eqn:LDA} is a research topic in its own right 
\cite{liu2004null,huang2002solving} so we won't explore it in more detail here. 

In addition to projecting the data into a reduced dimension space LDA can also be used for classification. In the LDA space each class $c \in C$ is modelled by a multivariate Gaussian distribution $P(\mathbf{X}|\mathbf{y} = c)$. So a query sample $x_i$ can be assigned to the class $ c_{LDA}$ for which the probability $P(y_i = c|x_i)$ is largest. This is effectively a Naive Bayes classifier:

\begin{equation}
\begin{split}
    c_{LDA}& = \arg\max_{c\in C}P(y_i = c|x_i) \\
    c_{LDA}& = \arg\max_{c\in C}P(x_i|y_i = c)P(y_i = c)
\end{split}
\end{equation}

In the code linked in the Appendix, LDA has 97\% accuracy on the hold-out set. Half the Penguins data is used to build the LDA model shown in Figure \ref{fig:PCA-LDA} and the other half is used for testing.

\section{Discussion \& Recommended Strategies}\label{sec:disc}

If the objective is to identify an effective feature selection strategy then there are many methods to choose from. 
Even though these methods have different inductive  biases we generally see good correlations between the feature rankings on the two datasets we consider. This shows that all these methods have some merit in identifying good features. Our objective is not to identify a single best strategy as this will depend to some extent on the data. Given this, we propose the following methodology for feature selection on a new dataset. 
\begin{enumerate}
    \item \textbf{Preliminary Analysis:} Use RF feature importance to rank all features (e.g. as shown in Figure \ref{fig:RF-FI}). This will provide some insight into what features are important and it may also indicate features that can be dropped from further consideration. We recommend the RF method as it considers features in context. If the objective of the exercise is to gain an insight into the data then the analysis may stop at this point. 
    \item \textbf{Subset Selection:} If the objective is to identify an effective subset for classification then a subset selection strategy is required. The main argument \emph{against} a Wrapper strategy is the computational cost. With advances in computing resources this is now less of an issue so we recommend a Wrapper strategy as described in Section \ref{sec:wrapper}. If the number of features still in consideration is not high then BE should be considered. If the set of possible features is large then SFS may be the pragmatic choice. The hybrid strategy described in Section \ref{sec:hybrid} could also be considered. 
\end{enumerate}

So our overall recommendation for feature subset selection is to use a Wrapper strategy. This would be in tune with the current model selection protocols in ML \cite{kumar2016model}. However, the recommendation for Preliminary Analysis is also important. Early in the analysis of new data it will be helpful to use a Filter strategy to gain some insight into the data. This Filter analysis will be independent of any specific classifier models that might be used subsequently.

\section{Conclusions}

The objective for this tutorial paper was to provide an overview of popular feature selection methods, describing how they work and providing links to Python implementations. It was not our intention to offer a comparative evaluation to identify the best methods; that is done very well elsewhere \cite{guyon2003introduction,molina2002feature}. Instead, our objective was to provide a stepping stone to help researchers get started with feature selection on their own datasets. 

In the Introduction we listed  objectives for feature selection other that to improve classifier performance. Indeed, the examples in this paper suggest that feature subset selection may not have a big impact on classifier performance - this is borne out by other studies \cite{guyon2003introduction}. Instead feature selection can deliver other benefits through savings in data gathering an model execution. Perhaps, the biggest benefit of feature selection  is the  insight it offers into what is important in the data under analysis.   

\section*{Acknowledgements}
This publication has emanated from research conducted with the financial support of Science Foundation Ireland under Grant numbers 16/RC/3872 and 18/CRT/6183. For the purpose of Open Access, the author has applied a CC BY public copyright licence to any Author Accepted Manuscript version arising from this submission.

\bibliographystyle{plain}  
\bibliography{FeatSel} 

\appendix
\section{Appendix I: Python Code}\label{app:code}

The GitHub repository\footnote{\url{https://github.com/PadraigC/FeatSelTutorial}} associated with this paper contains the following Python Notebooks:
\begin{itemize}
\item \textsf{FS-Wrappers}: Code for SFS and BE Wrappers from \textsf{mlxtend}.
\item \textsf{FS-Filters}: Code for using I-Gain and Chi-square Filters from \textsf{scikit-learn}.    
\item \textsf{FS-ReliefF}: Code for using ReliefF Filters from \textsf{skrebate}.
\item \textsf{FS-D-Tree}: Building D-Trees with embedded feature selction using  \textsf{scikit-learn}.
\item \textsf{FS-Lasso:} Feature selection for Logistic Regression using \textsf{scikit-learn}.
\item \textsf{FS-CFS:} Correlation-Based feature subset selection.
\item \textsf{FS-Random-Forest}: Feature importance from Random Forest using  \textsf{scikit-learn}.
\item \textsf{FS-PCA:} Principal Component Analysis using the PCA implementation in \textsf{scikit-learn}.
\item \textsf{FS-LDA:} Linear Discriminant  Analysis using the LDA implementation in \textsf{scikit-learn}.
\end{itemize}

\end{document}